\documentclass[unnumsec,webpdf,contemporary,large]{oup-authoring-template}%
\graphicspath{{Fig/}}

\usepackage{longtable}
\usepackage{url,multirow,tabulary}

\usepackage{subcaption}
\usepackage{booktabs}
\usepackage{makecell}
\usepackage{graphicx}
\usepackage{siunitx}
\hypersetup{colorlinks=true}

\newcommand{\eg}{\textit{e.g.}}
\newcommand{\etc}{\textit{etc}}
\newcommand{\etal}{\textit{et al.}}

\newcolumntype{L}[1]{>{\raggedright\arraybackslash}p{#1}}
\newcolumntype{C}[1]{>{\centering\arraybackslash}p{#1}}
\newcolumntype{R}[1]{>{\raggedleft\arraybackslash}p{#1}}

\begin{document}

\journaltitle{Transportation Safety and Environment}
\DOI{10.1093/tse/tdac026}
\copyrightyear{2022}
\pubyear{2022}
\appnotes{Review}

\firstpage{1}

\title[Computer Vision for Road Imaging and Pothole Detection: A State-of-the-Art Review of Systems and Algorithms]{\LARGE{Computer Vision for Road Imaging and Pothole Detection:\\A State-of-the-Art Review of Systems and Algorithms}}

\author[1]{{Nachuan Ma}\ORCID{0000-0002-7107-6577}}
\author[1]{{Jiahe Fan}\ORCID{0000-0002-2364-6811}}
\author[2]{{Wenshuo Wang}\ORCID{0000-0002-1860-8351}}
\author[3]{{Jin Wu}\ORCID{0000-0001-5930-4170}}
\author[4]{{Yu Jiang}\ORCID{0000-0001-8828-7832}}
\author[5]{{Lihua Xie}\ORCID{0000-0002-7137-4136}}
\author[1,$\ast$]{{Rui Fan}\ORCID{0000-0003-2593-6596}}

\authormark{Nachuan Ma \textit{et al}.}

\address[1]{Department of Control Science and Engineering, Tongji University, Shanghai 201804, P. R. China}
\address[2]{Department of Civil Engineering, McGill University, Montr\'eal, QC H3A 0C3, Canada}
\address[3]{Department of Electronics and Computer Engineering, the Hong Kong University of Science and Technology, Hong Kong SAR 999077, P. R. China}
\address[4]{CTO Office, ClearMotion Inc., Billerica, MA 01821, the United States,}
\address[5]{School of Electrical and Electronic Engineering, Nanyang Technological University, 50 Nanyang Avenue, Singapore 639798}

\corresp[$\ast$]{Corresponding author. \href{email:email-id.com}{rui.fan@ieee.org}}

\received{22}{12}{2021}
\revised{01}{04}{2022}
\accepted{12}{04}{2022}

\abstract{
Computer vision algorithms have been prevalently utilized for 3-D road imaging and pothole detection for over two decades. Nonetheless, there is a lack of systematic survey articles on state-of-the-art (SoTA) computer vision techniques, especially deep learning models, developed to tackle these problems. This article first introduces the sensing systems employed for 2-D and 3-D road data acquisition, including camera(s), laser scanners, and Microsoft Kinect. Afterward, it thoroughly and comprehensively reviews the SoTA computer vision algorithms, including (1) classical 2-D image processing, (2) 3-D point cloud modeling and segmentation, and (3) machine/deep learning, developed for road pothole detection. This article also discusses the existing challenges and future development trends of computer vision-based road pothole detection approaches: classical 2-D image processing-based and 3-D point cloud modeling and segmentation-based approaches have already become history; and Convolutional neural networks (CNNs) have demonstrated compelling road pothole detection results and are promising to break the bottleneck with the future advances in self/un-supervised learning for multi-modal semantic segmentation. We believe that this survey can serve as practical guidance for developing the next-generation road condition assessment systems.
}
\maketitle

\begin{figure*}[!t]
	\begin{center}
		\centering
		\includegraphics[width=0.99\textwidth]{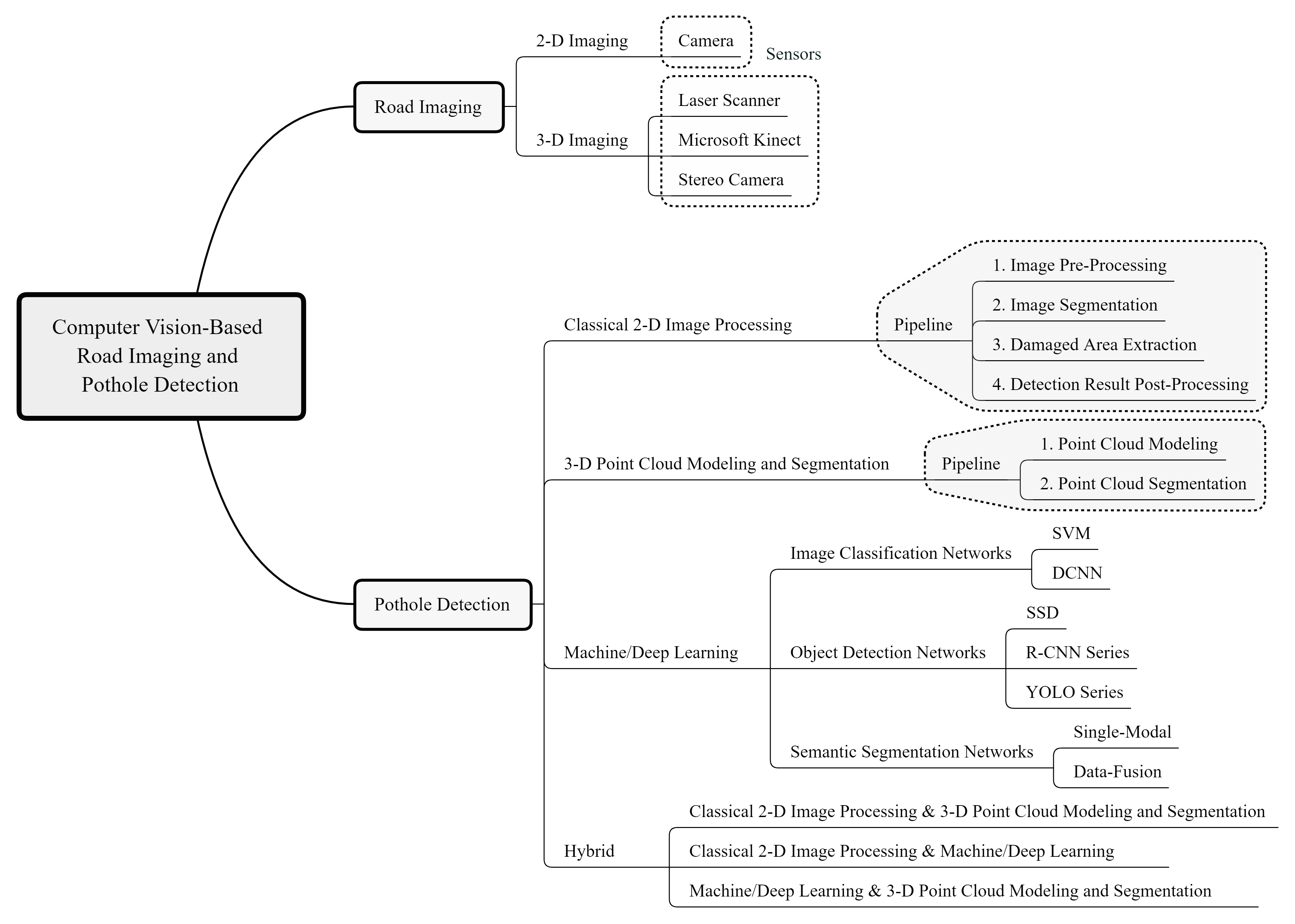}
		\centering
		\caption{An overview of road imaging systems and computer vision-based pothole detection algorithms.}
		\label{fig.mind_map}
	\end{center}
\end{figure*}

\section{Introduction}
A pothole is a considerably sizeable structural road failure \cite{mathavan2015review}. It is formed by the combined presence of water and traffic \cite{miller2003distress}. Water permeates the ground and weakens the soil under the road surface, and the traffic subsequently breaks the affected road surface, resulting in the removal of road surface chunks. 

Road potholes are not just an inconvenience, they are also a significant threat to vehicle conditions and traffic safety \cite{fan2019pothole}.
For instance, \textit{Chicago Sun-Times} reported that drivers filed 11,706 complaints about road potholes in the first two months of 2018 \cite{heaton_news}. According to \textit{The Pothole Facts}, approximately one-third of 33,000 traffic fatalities in the United States involve poor road conditions. 
It is, therefore, necessary and crucial to frequently inspect roads and repair potholes \cite{fan2021rethinking}.

Manual visual inspection is currently still the main form of road pothole detection \cite{fan2019road}. Structural engineers and certified inspectors regularly detect road potholes and report their locations. This process is inefficient, expensive, and dangerous. City councils in New Zealand, for example, spent millions of dollars in 2017 detecting and repairing potholes (Christchurch alone spent \$525K) \cite{guildford2018}. Additionally, it has been reported that more than 30K potholes are repaired in San Diego, the United States, each year. San Diego residents were suggested to report road potholes so as to relieve the burden of detection on the local road maintenance department \cite{Devine}. Further, manual road pothole detection results produced by inspectors and engineers are always subjective, as the decisions depend entirely on an individuals' experience and judgment \cite{koch2015review}. For these reasons, researchers have been dedicated to developing automated road condition assessment systems that can reconstruct, recognize, and localize road potholes efficiently, accurately, and objectively \cite{fan2021eusipco}. Specifically, in recent years, road pothole detection has become more than just an infrastructure maintenance problem because it is also a function of advanced driver-assistance systems (ADAS) embedded into L3/L4 self-driving cars by many automotive companies, and emerging autonomous driving systems call for a higher road maintenance standard \cite{fan2020we}. Jaguar Land Rover has experimented with data-driven technologies to inform drivers of pothole locations and issue warnings to slow down the car \cite{o2015jaquar}, while ClearMotion built an intelligent suspension system that uses a combination of hardware and software to anticipate, absorb, and counteract the shocks and vibrations caused by road potholes \cite{rosen2019}.

Since the turn of the millennium, computer vision techniques have been extensively employed to acquire 3-D road data and/or detect road potholes. However, the latest survey on this research topic rarely discusses cutting-edge computer vision techniques, such as 3-D point cloud modeling and segmentation, machine/deep learning, \etc. This article provides a comprehensive and thorough review of the state-of-the-art (SoTA) road imaging systems and computer vision-based pothole detection algorithms. An overview of the existing systems and algorithms is shown in Fig. \ref{fig.mind_map}. Laser scanners, Microsoft Kinect sensors, and camera(s) are the three most prevalently used sensors for road data acquisition. The existing road pothole detection approaches are categorized into four groups: (1) classical 2-D image processing-based \cite{koch2011pothole}, (2) 3-D point cloud modeling and segmentation-based \cite{chang2005detection}, (3) machine/deep learning-based \cite{lin2010potholes}, and (4) hybrid \cite{fan2019pothole}. This article  systematically reviews the prior arts (see Sec. \ref{sec.road_imaging_systems} and Sec. \ref{sec.road_pothole_detection}) and the open-access datasets (see Sec. \ref{sec.public_datases}), and discusses the existing challenges and their possible solutions (see Sec. \ref{sec.existing_challenges}). We believe that this article can provide readers with guidance when developing the next-generation 3-D road imaging and pothole detection algorithms. 

\begin{figure*}[!t]
	\begin{center}
		\centering
		\includegraphics[width=0.80\textwidth]{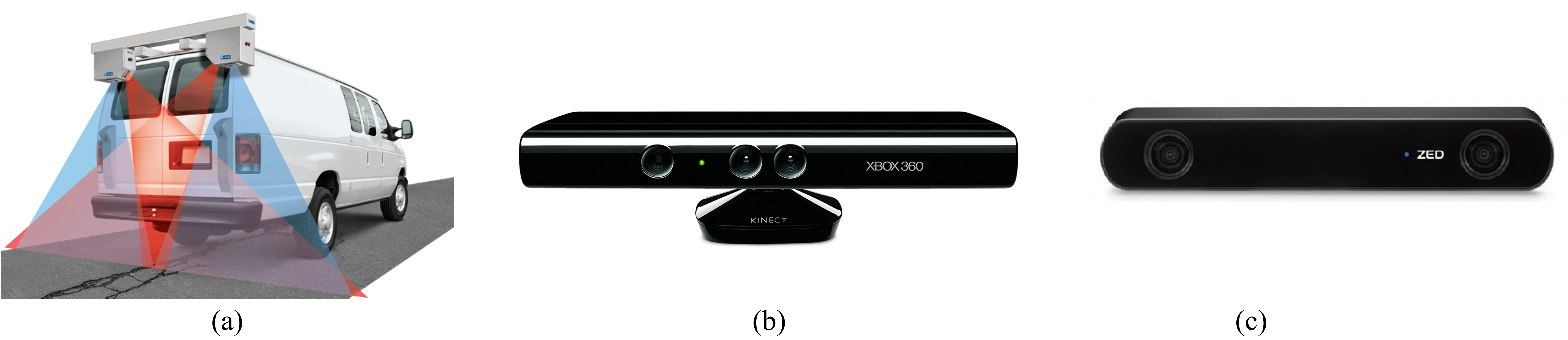}
		\centering
		\caption{Commonly used sensors for 3-D road data acquisition: (a) laser scanner \cite{laurent2012using}; (b) Microsoft Kinect \cite{banerjee2017exploratory}; (c) stereo camera \cite{Stereo}.}
		\label{fig.sensors}
	\end{center}
\end{figure*}

\section{Road Imaging Systems}\label{sec.road_imaging_systems}

Road imaging (or road data acquisition) is typically the first step of intelligent road inspection \cite{fan2021eusipco}. Cameras and range sensors have been extensively used to acquire visual road data. The use of 2-D imaging technology for this task began as early as 1991 \cite{mahler1991pavement}. However, the geometric structure of a road surface cannot be illustrated from unrelated 2-D road images (without overlapping areas) \cite{fan2018road}. Additionally, the image segmentation algorithms performing on either gray-scale or color road images can be severely affected by various environmental factors, most notably by poor illumination conditions \cite{jahanshahi2013unsupervised}. Many researchers \cite{fan2021rethinking,fan2018road,fan2019real,fan2018real} have thus resorted to 3-D imaging technologies, which are more feasible to overcome these two drawbacks. The most commonly used sensors for 3-D road data acquisition include laser scanners, Microsoft Kinect sensors, and stereo cameras, as shown in Fig. \ref{fig.sensors}.

Laser scanning is a well-established imaging technology for accurate 3-D road data acquisition \cite{mathavan2015review}. This technology is developed based on trigonometric triangulation \cite{laurent1997road}. The sensor (receiver) is located at a position with a known distance from the laser illumination source \cite{fan2021bookchapter}. Accurate point measurements can, therefore, be made by calculating the reflection angle of the laser light. However, laser scanners have to be mounted on specific road inspection vehicles (see Fig. \ref{fig.sensors}(a)) \cite{tsai2018pothole} for 3-D road data acquisition. Such vehicles are not widely used because of high equipment purchase and long-term maintenance costs.
 
Microsoft Kinect sensors \cite{jahanshahi2013unsupervised} were initially designed for the Xbox-360 motion-sensing games, and are typically equipped with an RGB camera, an infrared sensor/camera, an infrared emitter, microphones, accelerometers, and a tilt motor for motion tracking  \cite{mathavan2015review}. There have been three reported efforts \cite{jahanshahi2013unsupervised,tsai2018pothole,moazzam2013metrology} on 3-D road data acquisition using Microsoft Kinect sensors. Although such sensors are cost-effective and convenient to use, they greatly suffer from infra-red saturation in direct sunlight, and the 3-D road surface reconstruction accuracy is unsatisfactory \cite{fan2019pothole}. 

\begin{figure}[!t]
	\begin{center}
		\centering
		\includegraphics[width=0.485\textwidth]{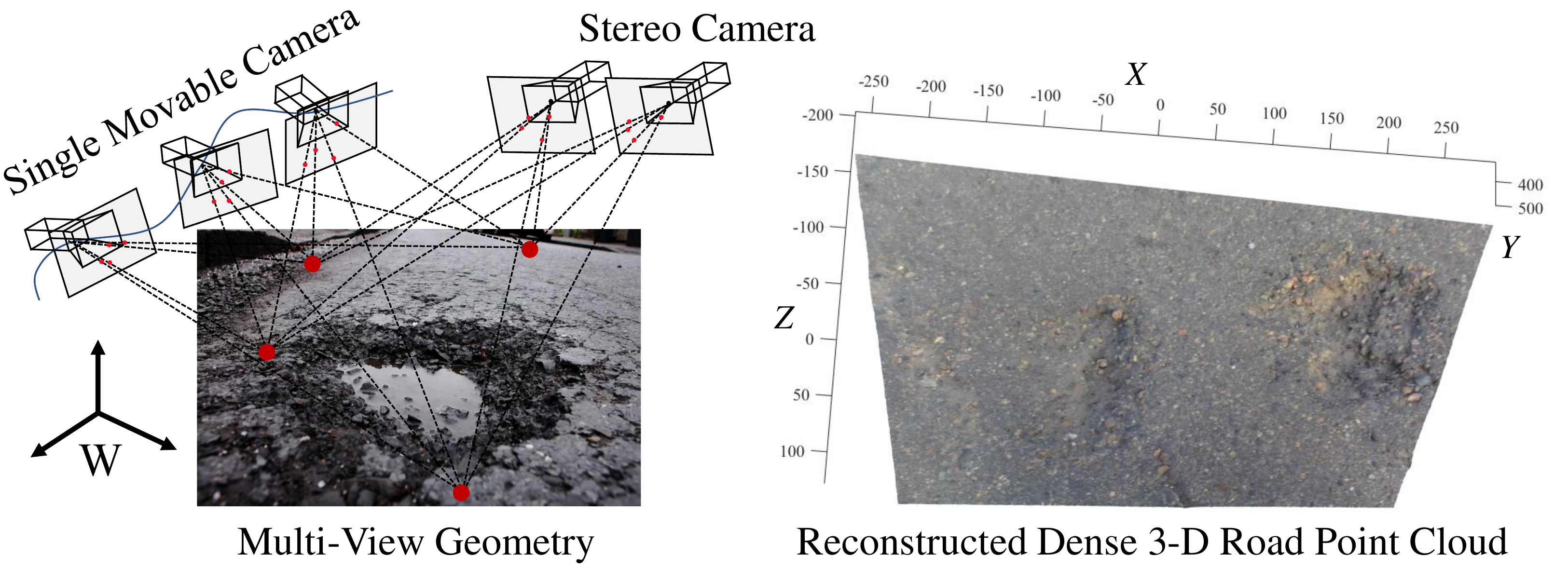}
		\centering
		\caption{3-D road imaging with camera(s).}
		\label{fig.road_imaging}
	\end{center}
\end{figure}

3-D road data can also be obtained using multiple 2-D road images captured from different views, \eg, using either a single movable camera \cite{jog2012pothole} or an array of synchronized cameras \cite{fan2019real}, as illustrated in Fig. \ref{fig.road_imaging}. The theory behind this technique is generally known as \textit{multi-view geometry} \cite{hartley2003multiple}. The essential task of 3-D geometry reconstruction from multiple views is sparse or dense correspondence matching. A typical monocular sparse road surface 3-D reconstruction approach, as shown in \cite{zhang2012unmanned}, where the camera poses and sparse 3-D road point clouds are obtained using a structure from motion (SfM) \cite{ullman1979interpretation} algorithm and are refined using a bundle adjustment (BA) \cite{triggs1999bundle} algorithm. 

\begin{figure}[!t]
	\begin{center}
		\centering
		\includegraphics[width=0.48\textwidth]{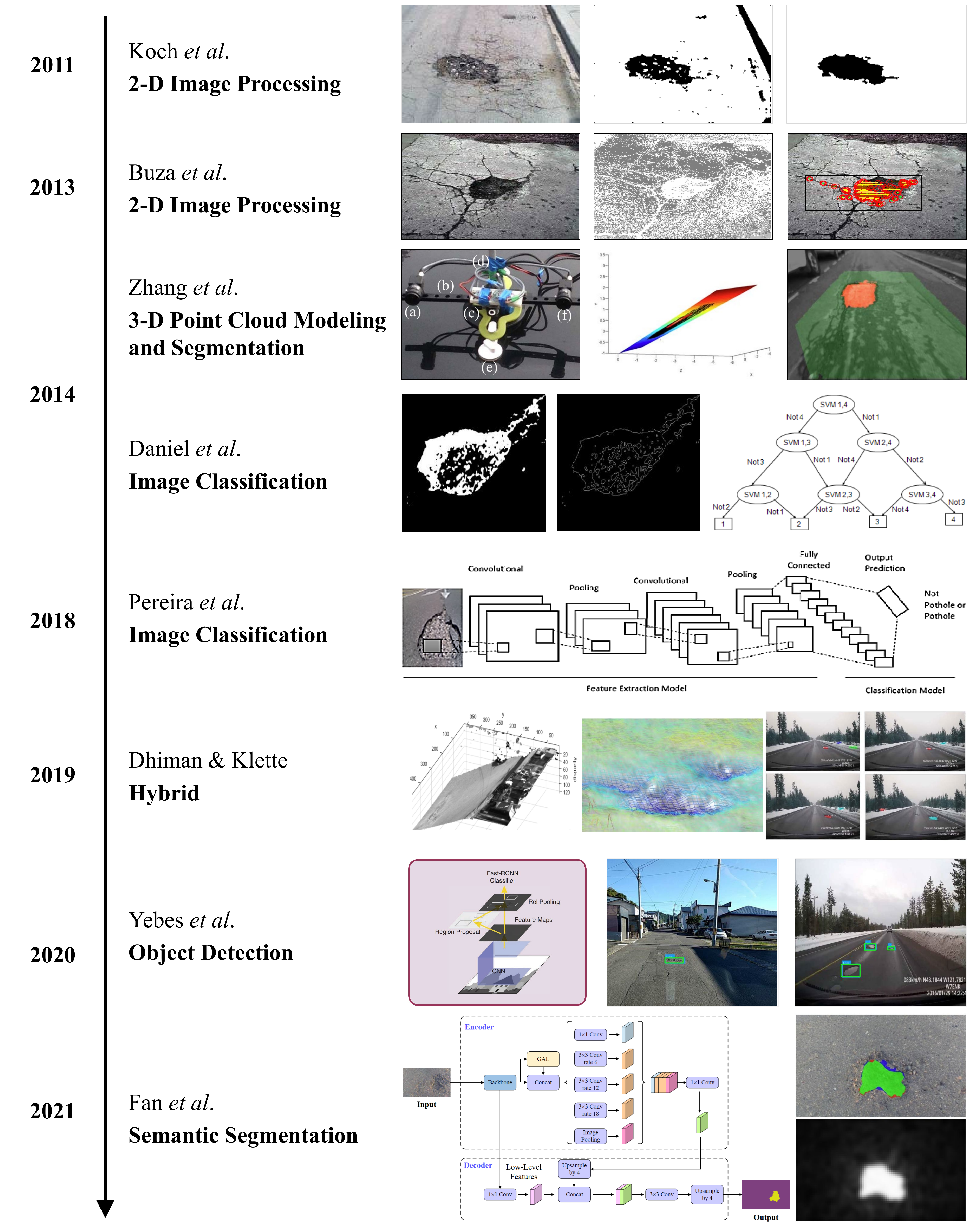}
		\centering
		\caption{
			The most representative road pothole detection algorithms developed between 2011 and 2021.
		}
		\label{fig.key_researches}
	\end{center}
\end{figure}

 The use of stereo cameras for dense 3-D road point cloud acquisition was pioneered by researchers \cite{fan2018road, zhang2013advanced, ozgunalp2016vision} from Bristol Visual Information Laboratory. In this case, depth information is acquired by finding the horizontal positional differences (disparities) of the visual feature correspondence pairs between two synchronously captured road images \cite{fan2020computer}. This process is commonly referred to as \textit{disparity estimation} or \textit{stereo matching}, which mimics human binocular vision. \cite{zhang2013advanced} proposed a seed-and-grow disparity estimation algorithm to acquire 3-D road data efficiently. \cite{ozgunalp2016vision} introduced a more adaptive disparity search range propagation strategy to improve the accuracy of the estimated road disparities. \cite{fan2021rethinking,fan2018road} utilized a perspective transformation algorithm to transform the target image into the reference view, which significantly minimizes the trade-off between stereo matching speed and disparity accuracy.
Additionally, the bottleneck problems existing in \cite{zhang2013advanced} and \cite{ozgunalp2016vision} were also tackled with the use of efficient and adaptive cost volume processing algorithms. It is reported in \cite{fan2021rethinking} and \cite{fan2018road} that the accuracy of the reconstructed 3-D road geometry models is over 3 mm (an example is given in Fig. \ref{fig.road_imaging}). Compared to laser scanners and Microsoft Kinect sensors, stereo cameras are cheaper and more reliable for 3-D road imaging. With the recent advances in deep learning, convolutional neural networks (CNNs) have demonstrated greater disparity estimation results than traditional explicit programming methods. Their limitations and future development trends will be discussed in Sec. \ref{sec.existing_challenges}.


\begin{table*}[!t]
	\renewcommand{\arraystretch}{1.3}
	\caption{Representative classical 2-D image processing-based approaches.}
	\label{tab.2d_image_processing}
	\centering
		
		\centering
		\begin{tabular}{C{2.39cm}C{2.4cm}p{11.45cm}}
			\toprule
			Reference & Input & Details \\
			\hline
			
			Koch and Brilakis \cite{koch2011pothole} (2011) & Color image &
			A road image is segmented into damaged and undamaged road regions using a histogram-based thresholding method. The damaged road areas are processed with morphological operations and elliptic regression. The road potholes are detected by comparing the road textures inside and outside the ellipse.
			\\
			
			Buza  \etal \ \cite{buza2013pothole} (2013)
			& Color image & 
			Otsu's thresholding method is adopted to segment road images. Spectral clustering is utilized to extract damaged road areas (potholes). 
			\\
			
			Ryu \etal \ \cite{ryu2015feature} (2015) 
			& Color image & 
			Road images are processed with morphological filters and segmented using a histogram-based thresholding method. A potential road pothole contour is extracted based on geometric properties. An ordered histogram intersection method is used to determine whether the extracted area contains a road pothole. 
			\\
			
			Schiopu {\etal} \cite{schiopu2016pothole} (2016) & Color image & 
			A histogram-based thresholding method is utilized to generate a set of road pothole candidates. The candidates with specific geometric properties are determined to be road potholes. 
			\\
			
			Jak{\v{s}}tys {\etal} \cite{jakvstys2016detection} (2016) & Color image & 
			Triangle thresholding and adaptive thresholding methods are used to segment road images. A heuristic edge detection approach is designed for road pothole contour extraction.
			\\
			
			Akagic {\etal} \cite{akagic2017pothole} (2017) & Color image
			& 
			Road pothole regions of interest (RoIs) are detected by (1) manipulating the B component in the RGB color space and (2) performing two-level dynamic road pixel selection. The search for road potholes is conducted only in the RoIs. The road potholes are detected by comparing two cropped road images based on the method proposed in \cite{buza2013pothole}.
			\\
			
			Wang {\etal} \ \cite{wang2017asphalt} (2017) & Gray-scale image & 
			The wavelet energy field of a road image is constructed to highlight road potholes. Damaged road areas are processed with morphological filters. A Markov random fields-based image segmentation method is used to segment the damaged road areas for pothole detection. Morphological filters are used again to refine the road pothole detection results. 
			\\
			
			Chung and Khan \cite{chung2019watershed} (2019) & Gray-scale image
			&
			Otsu’s thresholding method is used to segment road images. The segmented images are processed with morphological filters before performing distance transform. The watershed algorithm is applied to the distance transform images for road pothole detection. 
			\\
			Moazzam {\etal} \cite{moazzam2013metrology} (2013) & Depth images &
			The road potholes are detected by analyzing road depth distribution w.r.t. different azimuth and elevation angles. The approximate volume of each road pothole is calculated using the trapezoidal rule with unit spacing on the area-depth curves. 
			\\		
			Fan {\etal} \cite{fan2019road} (2019) & Transformed disparity image &
			A dense road disparity image is transformed to better distinguish the damaged and undamaged road areas. The transformed disparity image is segmented using Otus’s thresholding method for road pothole detection.
			\\
			Fan {\etal} \cite{fan2021rethinking} (2021) & Transformed disparity image & 
			SLIC is utilized to group the transformed disparities into a collection of superpixels. The road potholes are then detected by finding the superpixels, whose values are lower than an adaptively determined threshold. 
			\\
			
			\bottomrule
		\end{tabular}
	\end{table*}

\section{Road Pothole Detection Approaches}\label{sec.road_pothole_detection}
The taxonomy of SoTA computer vision-based road pothole detection algorithms is illustrated in Fig. \ref{fig.mind_map}. The classical 2-D image processing-based algorithms process (\eg, enhance, compress, transform, segment) road RGB or disparity/depth images with explicit programming \cite{koch2015review}. Machine/deep learning-based algorithms address the road pothole detection problem using image classification, object recognition, or semantic segmentation algorithms, solvable with SoTA CNNs \cite{fan2021graph}. 3-D road point cloud modeling and segmentation-based algorithms fit a specific geometry model (typically a planar or quadratic surface) to the observed road point cloud and segment the road point cloud by comparing the observed and fitted surfaces \cite{fan2019pothole}. Hybrid methods combine two or more categories of algorithms mentioned above to improve the overall road pothole detection performance. The most representative road pothole detection algorithms (from classical 2-D image processing-based to deep learning-based) developed between 2011 and 2021 are shown in Fig. \ref{fig.key_researches}.

\subsection{Classical 2-D Image Processing}\label{sec.2d-algorithms}

Classical 2-D image processing-based road pothole detection is a well-researched topic. As shown in Fig. \ref{fig.mind_map}, such approaches generally have a four-stage pipeline: (1) image pre-processing, (2) image segmentation, (3) damaged area extraction, and (4) detection result post-processing \cite{koch2015review}. The representative prior arts are summarized in Table \ref{tab.2d_image_processing}.

Image pre-processing algorithms, such as median filtering \cite{wang2017asphalt}, Gaussian filtering \cite{kim2014system}, bilateral filtering \cite{fan2019roadcrack}, and morphological filtering \cite{pitas2000digital}, are first utilized to reduce redundant information and highlight the damaged road areas. For instance, an adaptive histogram equalization algorithm is used in \cite{kim2014system} to adjust the image brightness before binarizing the road images, and a Leung-Malik filter \cite{leung2001representing} and Schmid filter \cite{schmid2001constructing} are used in \cite{koch2011pothole} to emphasize structural texture characteristics in color road images. Recently, many researchers \cite{fan2019pothole,fan2021rethinking,fan2019road,moazzam2013metrology,fan2018novel} have resorted to 2-D spatial visual information (typically road depth/disparity images) for pothole detection. For example,  \cite{fan2018novel} and \cite{fan2019pothole} transformed a road disparity image with a stereo rig roll angle and road disparity projection model, which were estimated by minimizing a global energy function using golden section search \cite{pierre1986optimization} and dynamic programming \cite{ozgunalp2016multiple} algorithms. Disparity transformation makes the damaged road areas highly distinguishable, as illustrated in Fig. \ref{fig.GAL-Deeplabv3+}. \cite{fan2019road} yields the closed-form solution of the above energy minimization problem, and therefore, avoids the intensive computations in the iterative optimization process. As depth/disparity images can depict the geometric structure of road surfaces, they are more informative for pothole detection \cite{fan2019road}. 

The pre-processed road images are then segmented to separate foreground (damaged road areas) and background (undamaged road areas). Most prior arts \cite{fan2019roadcrack,jakvstys2016detection,buza2013pothole} employ histogram-based thresholding methods, such as Otsu’s thresholding \cite{otsu1979threshold}, triangle thresholding \cite{koch2011pothole}, and adaptive thresholding \cite{fan2019roadcrack,jakvstys2016detection}, to segment color/gray-scale road images. As discussed in \cite{buza2013pothole}, Otsu’s thresholding \cite{otsu1979threshold} method minimizes the intra-class variance and achieves better performance than the triangle thresholding \cite{koch2011pothole} method in terms of segmenting road images.  \cite{jakvstys2016detection} employs an adaptive thresholding method to segment road images, and it also outperforms the commonly used triangle thresholding method. Recent works \cite{fan2019pothole,fan2021rethinking,fan2019road,fan2018novel} demonstrated that such image segmentation algorithms typically work more effectively and accurately on transformed disparity images, depicting the quasi bird's eye view of the road scene. For example, \cite{fan2019pothole} utilizes Otsu’s thresholding \cite{otsu1979threshold} method to segment the transformed disparity images for road pothole detection, and in \cite{fan2021rethinking}, a simple linear iterative clustering (SLIC) algorithm \cite{achanta2012slic} is used to group the transformed disparities into a collection of superpixels. The road potholes are then detected by finding the superpixels, whose values are lower than an adaptively determined threshold.

\begin{table*}[!t]
	\renewcommand{\arraystretch}{1.3}
	\centering
	\caption{Representative 3-D point cloud modeling and segmentation-based approaches.}  
	\begin{tabular}{C{2.9cm}C{2.2cm}C{3.8cm}p{7.1cm}} 
		\toprule
		Reference & Input & Key algorithm(s) & Details \\
		\hline
		Zhang and Elaksher \cite{zhang2012unmanned} (2012) & 3-D point cloud & SfM, BA, 3-D feature extraction &
		Sparse 3-D road geometry models are reconstructed with SfM and refined with BA. Road potholes are detected by finding distinguishable 3-D features.
		\\
		
		Zhang \cite{zhang2013advanced} (2013) & 3-D point cloud & Stereo vision, quadratic surface fitting, connected component labeling (CCL) & 
		A quadratic surface is fitted to the observed 3-D road point cloud. The 3-D points under the fitted surface are considered part of road potholes. Different road potholes are labeled using CCL.
		\\
		
		Li {\etal} \cite{li2018road} (2018) & 3-D point cloud & Stereo vision, planar surface fitting, bi-square weighted robust least-squares approximation, CCL &
		An observed 3-D road point cloud is interpolated into a planar surface using a bi-square weighted robust least-squares approximation. The 3-D points under the fitted surface are considered to be part of road potholes. CCL is also used to label different road potholes.
		\\
		
		Du {\etal} \cite{du2020pothole} (2020) & 3-D point cloud & Stereo vision, planar surface fitting and segmentation, K-means clustering, region growing & 
		
		The surface normal information is incorporated into the road surface modeling process. K-means clustering and region growing algorithms are used to extract road potholes.
		\\
		
		\bottomrule
	\end{tabular}
	\label{tab.3d_pc_based_methods}
\end{table*}

The third and fourth stages are typically performed in a joint manner. The damaged road areas (potholes) are first extracted from the segmented foreground based on geometric and textural assumptions \cite{fan2021rethinking}:
\begin{itemize}
\item Potholes are typically concave holes;
\item The pothole texture is typically grainier and coarser than that of the surrounding road surface;
\item The intensities of the pothole RoI pixels are typically lower than those of the surrounding road surface due to shadows.
\end{itemize}
For example, in \cite{koch2011pothole}, the contour of a potential pothole is modeled as an ellipse. The image texture within the ellipse is then compared with that of the undamaged road areas. If the elliptical RoI has a coarser and grainier texture than that of the surrounding region, the ellipse is identified as a road pothole. In \cite{ryu2015feature}, the contour of a potential pothole is extracted by analyzing various geometric features, such as size, compactness, ellipticity, and convex hull. An ordered histogram intersection method is then used to determine whether the extracted region contains a road pothole. Finally, the extracted damaged road areas are post-processed to further improve the road pothole detection results. This process is typically similar to the first stage.  

Classical 2-D image processing-based road pothole detection approaches have been researched for almost two decades. This type of algorithms have been systematically studied by \cite{koch2015review} and we refer readers to \cite{koch2015review} for more details. However, such approaches were developed based on early techniques and can be severely affected by various environmental factors. Fortunately, modern 3-D computer vision and machine learning algorithms have greatly overcome these shortcomings.

\begin{table*}[!t]
	\renewcommand{\arraystretch}{1.3}
	\caption{Image classification-based approaches.}
	\centering
	\begin{tabular}{C{2.75cm}C{2.4cm}C{2.8cm}p{8.0cm}}
		\toprule
		Reference & Input  & Key algorithm(s) & Details \\
		\hline
		
		Lin and Liu \cite{lin2010potholes} (2010) & Gray-scale image & NL-SVM & 
		Average gray level, contrast, consistency, entropy, and three-order moments of gray-scale road images are computed to create hand-crafted visual features; An NL-SVM model is trained to learn these features for road image classification.
		\\
		
		Daniel and Preeja \cite{daniel2014automatic} (2014) & Gray-scale image
		& SVM &
		Classical image processing algorithms are utilized to reduce road image noise and highlight informative visual features; CCL is then employed to obtain the connected components; The five most prominent components are selected as training samples to train an SVM model for road image classification.
		\\
		
		Hadjidemetriou {\etal} \cite{hadjidemetriou2016automated} (2016) 
		& Gray-scale image
		& SVM, DCT, GLCM &
		Road image patches are utilized to generate feature vectors using discrete cosine transform (DCT) \cite{ahmed1974discrete} and gray-level co-occurrence matrix (GLCM) algorithms \cite{haralick1973textural}. An SVM model is then trained with such feature vectors to realize binary road patch classification. 
		\\
		
		Hoang \cite{hoang2018artificial} (2018) & Gray-scale image
		& LS-SVM, ANN &
		Classical image processing algorithms are used to generate hand-crafted visual features; A least-squares SVM (LS-SVM) model and an artificial neural network (ANN) model are trained with such hand-crafted visual features to recognize road images containing potholes.
		\\
		
		Pan {\etal} \cite{pan2018detection} (2018) & Color image, Multi-spectral image
		& ANN, RF, SVM &
		Spectral, geometric, and textural features are extracted; Three models: ANN, random forest (RF), and SVM, are trained to learn these features for road image classification.
		\\
		
		Gao {\etal} \cite{gao2020detection} (2020) & Color image
		& LIBSVM & 
		Classical image processing algorithms, including binarization, morphology operations, and integral projection, are used to generate hand-crafted visual features; A model based on the library for SVM (LIBSVM) is trained to detect road potholes and cracks.
		\\
		
		Pereira {\etal} \cite{pereira2018deep} (2018) & Color image & Self-designed DCNN &
		A DCNN, consisting of four convolutional-pooling layers and one FC layer, is developed from scratch to classify road images. 
		\\
		
		An {\etal} \cite{an2018detecting} (2018) & Color image, gray-scale image
		& Inception, ResNet, and MobileNet & 
		Four existing DCNNs are trained to classify color and gray-scale road image patches.
		\\
		
		Ye {\etal} \cite{ye2021convolutional} (2019) & Color image
		& Self-designed DCNN &
		A DCNN containing a pre-pooling layer (used to reduce the characteristics unrelated to road potholes) is designed from scratch to classify road images. 
		\\
		
		Bhatia {\etal} \cite{bhatia2019convolutional} (2019) & Thermal image
		& Self-designed DCNN & 
		A DCNN model (with ResNet as the backbone network) is designed to classify thermal road images.
		\\
		
		\bottomrule
	\end{tabular}
	\label{tab.image_classification}
\end{table*}

\subsection{3-D Point Cloud Modeling and Segmentation}\label{sec.3d_modeling_algorithms}

An example of the reconstructed dense 3-D road point clouds is given in Fig. \ref{fig.road_imaging}. 
The approaches designed to process 3-D road point clouds generally have a two-stage pipeline \cite{zhang2013advanced, wu2021ist}: (1) interpolating the observed 3-D road point cloud into an explicit geometric model (typically a planar or quadratic surface), and (2) segmenting the observed 3-D road point cloud by comparing it with the interpolated geometric model. The most representative algorithms are summarized in Table \ref{tab.3d_pc_based_methods}.

Taking \cite{zhang2013advanced} as an example, quadratic surfaces are fitted to dense 3-D road point clouds using least-squares fitting. By comparing the difference (elevation) between the observed and fitted 3-D road surfaces, the damaged road areas (potholes) can be effectively extracted. Different potholes are also labeled using a connected component labeling (CCL) algorithm. Similarly, \cite{du2020pothole} interpolates the observed 3-D road point clouds into planar surfaces. The potential road potholes are roughly detected by finding the 3-D points under the fitted surface. K-means clustering \cite{hartigan1979algorithm} and region growing algorithms are subsequently used to refine the road pothole detection results. 

Least-squares fitting, however, can be severely affected by outliers, often making the modeled road surface inaccurate \cite{fan2019pothole}. Therefore, \cite{li2018road} employs the bi-square weighted robust least-squares approximation for road point cloud modeling. \cite{fan2018novel} utilized the random sample consensus (RANSAC) algorithm \cite{fischler1981random} to improve the robustness of quadratic surface fitting. \cite{ozgunalp2016vision} and \cite{fan2019pothole} incorporated surface normal information into the process of quadratic surface fitting, which greatly enhances the performance of freespace and road pothole detection.

In addition to the aforementioned camera-based approaches, \cite{li2009real} employs high-speed 3-D transverse scanning technology for road shoving (abrupt waves across the road surface) and pothole detection. A subpixel line extraction method (including point cloud filtering, edge detection, and spline interpolation) is performed on the laser stripe data. The road transverse profile is then generated from the laser stripe curve and is approximated by line segments. The second-order derivatives of the segment endpoints are used to identify the feature points of possible shoving and potholes. Recently, \cite{ravi2020highway} introduced a LiDAR-based road pothole detection system, where the 3-D road points are classified as damaged and undamaged by comparing their distances to the best-fitting planar 3-D road surface. Unfortunately, \cite{ravi2020highway} lacks the algorithm details and necessary quantitative experimental road damage detection results. 

3-D point cloud modeling and segmentation-based methods are relatively rare compared to other approaches.
Nevertheless, actual roads are always uneven, making such approaches occasionally infeasible. Furthermore, acquiring 3-D road point clouds might not be necessary if the objective is only to recognize and localize road potholes instead of obtaining their geometric details. With the combination of 2-D image processing algorithms, the 3-D point cloud modeling performance can be significantly boosted \cite{fan2019pothole}.

\subsection{Machine/Deep Learning}\label{sec.machine_deep_learning}

With recent advances in machine/deep learning, deep CNNs (DCNNs) have become the mainstream techniques for road pothole detection. Instead of setting explicit parameters to segment road images or point clouds for pothole detection, DCNNs are typically trained through back-propagation with a large amount of human-annotated road data \cite{lecun2015deep}. Data-driven road pothole detection approaches are generally developed based on three types of techniques \cite{fan2021bookchapter}: (1) image classification networks, (2) object detection networks, and (3) semantic segmentation networks. Image classification networks are trained to classify positive (pothole) and negative (non-pothole) road images, object detection networks are trained to recognize road potholes at the instance level, and semantic segmentation networks are trained to segment road (color or disparity/depth) images for pixel-level (or semantic-level) road pothole detection. The remainder of this section details each type of these algorithms.

\begin{figure*}[!t]
	\begin{center}
		\centering
		\includegraphics[width=0.99\textwidth]{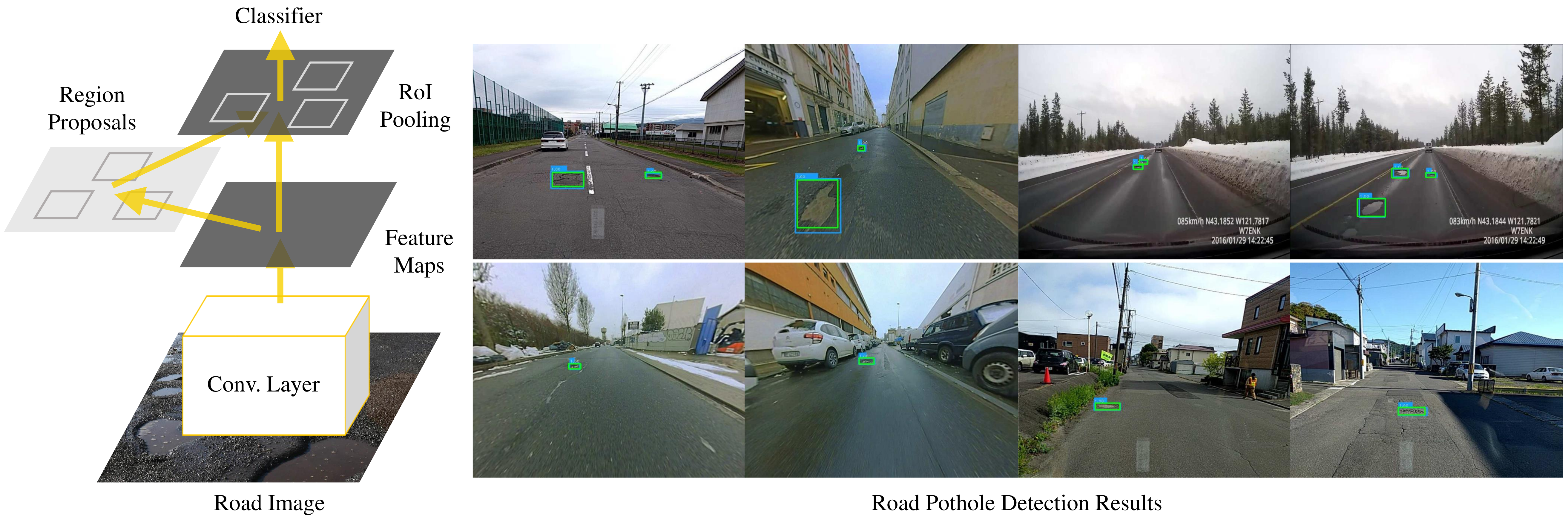}
		\centering
		\caption{Faster R-CNN architecture \cite{ren2015faster} and road pothole detection results \cite{yebes2020learning}. }
		\label{fig.faster_r-cnn}
	\end{center}
\end{figure*}

\subsubsection{Image Classification-Based Methods}
Before deep learning technology exploded, researchers typically used classical image processing algorithms to generate hand-crafted visual features and trained a support vector machine (SVM) \cite{cortes1995support} model to classify road image patches. The most representative SVM-based approaches \cite{lin2010potholes,daniel2014automatic,hadjidemetriou2016automated,pan2017object,pan2018detection,hoang2018artificial,gao2020detection} are summarized in Table \ref{tab.image_classification}. As such algorithms are already outdated, we do not present readers with too many details here.

With the revolution of computational resources and the increase in training data sample size, DCNNs have been extensively used for road pothole detection. Compared to the traditional SVM-based approaches, DCNNs are capable of learning more abstract (hierarchical) visual features, and they have significantly improved the road pothole detection performance \cite{fan2019roadcrack}. The most typical DCNN-based approaches \cite{pereira2018deep, ye2021convolutional, bhatia2019convolutional,an2018detecting} are summarized in Table \ref{tab.image_classification}. \cite{pereira2018deep} and \cite{ye2021convolutional} designed DCNNs from scratch. The DCNN presented in \cite{pereira2018deep} consists of four convolutional-pooling layers and one fully connected (FC) layer. Extensive experiments on the road data collected in Timor-Leste demonstrated the effectiveness of such a DCNN in terms of classifying pothole and non-pothole images. The DCNN introduced in \cite{ye2021convolutional} consists of a pre-pooling layer, three convolutional-pooling layers, a sigmoid layer, and two FC layers. The pre-pooling layer was designed to reduce the characteristics unrelated to road potholes. The experimental results suggests that the proposed pre-pooling layer can greatly improve the performance of road image classification, and that the designed DCNN can effectively detect road potholes under different illumination conditions.

 \cite{bhatia2019convolutional} and \cite{an2018detecting} developed road image classification networks based on existing DCNNs. \cite{bhatia2019convolutional} developed a DCNN based on the popular residual network \cite{he2016deep}. Extensive experiments demonstrated that the proposed model can effectively classify thermal road images collected in the night and/or foggy weather, and it also outperforms the prior arts \cite{hoang2018artificial, ryu2015image, an2018detecting}. In \cite{an2018detecting}, four well-developed DCNNs: (1) Inception-v4 \cite{szegedy2017inception}, (2) Inception with ResNet-v2 \cite{szegedy2017inception}, (3) ResNet-v2 \cite{he2016identity}, and (4) MobileNet-v1 \cite{howard2017mobilenets}, are trained to classify road images. The experimental results suggests that these models performed similarly on the test set. Recently, \cite{fan2021ist} compared 30 SoTA image classification DCNNs in terms of detecting road cracks and found that road crack detection is a relatively easy task compared to the image classification tasks in other application domains. Road pothole detection is an easier task compared to road crack detection. Therefore, we believe that road pothole detection with image classification networks is a well-solved problem.

\subsubsection{Object Detection-Based Methods}

As illustrated in Fig. \ref{fig.mind_map}, object detection-based road pothole detection approaches can be grouped into three types: (1) single shot multi-box detector (SSD)-based, (2) region-based CNN (R-CNN) series-based, and (3) you only look once (YOLO) series-based. The most representative object detection-based approaches are summarized in Table \ref{tab.object_detection_methods}. 

An SSD has two components \cite{liu2016ssd}, namely a backbone model and an SSD head. The former is a deep image classification network for visual feature extraction, while the latter is one or more convolutional layers added to the backbone so that the outputs can be bounding boxes with object classes. Researchers in this field have mainly incorporated different image classification networks into the SSD for road pothole detection. For example, Inception-v2 \cite{szegedy2016rethinking} and MobileNet \cite{howard2017mobilenets} were used as the backbone networks in \cite{maeda2018road}, while ResNet-34 \cite{he2016deep} and RetinaNet \cite{lin2017focal} were used as the backbone networks in \cite{gupta2020detection}.

\begin{table*}[!t]
	\renewcommand{\arraystretch}{1.3} 
	\centering
	\caption{Object detection-based approaches.}
	\begin{tabular}{C{2.4cm}C{2.3cm}C{2.6cm}p{8.7cm}}
		\toprule
		Reference & Input  & Key algorithm(s) & Details \\
		\hline
		Suong {\etal} \cite{suong2018detection} (2018) & Color image & YOLO & 
		Two object detection networks: F2-Anchor and Den-F2-Anchor, developed based on YOLOv2, are trained to detect potholes in the color road images.
		\\
		Maeda {\etal} \cite{maeda2018road} (2018) & Color image
		& SSD & 
		Two SSD-based DCNNs (with Inception-v2 and MobileNet as the backbone networks, respectively) are trained to detect potholes in color road images.
		\\
		
		Wang {\etal} \cite{wang2018road} (2018) & Color image &
		Faster R-CNN &
		Two Faster R-CNNs (with ResNet-101 and ResNet-152 as the backbone networks, separately) are trained to detect road potholes.
		\\
		
		Ukhwah {\etal} \cite{ukhwah2019asphalt} (2019) & Gray-scale image
		& YOLO & 
		YOLOv3, YOLOv3 Tiny, and YOLOv3 SPP are trained to detect potholes in gray-scale road images. YOLOv3 SPP achieves the best overall performance.
		\\
		
		Dharneeshkar {\etal} \cite{dharneeshkar2020deep} (2020) 
		& Color image
		& YOLO & 
		YOLOv2, YOLOv3, and YOLOv3 Tiny are trained to detect road potholes. YOLOv3 Tiny achieves the highest mAP, precision, and recall.
		\\
		
		Baek and Chung \cite{baek2020pothole} (2020) & Color image
		& YOLO & 
		Two YOLOv1 models are trained to detect cars (background) and road potholes (in the foreground).
		\\
		Kortmann {\etal} \cite{kortmann2020detecting} (2020) & Color image & Faster R-CNN &
		A classifier is first trained to infer the country where the road image was taken. A Faster R-CNN is then trained w.r.t. each country for road crack and pothole detection.
		\\
		Yebes {\etal} \cite{yebes2020learning} (2020) & Color image
		& 
		Faster R-CNN, SSD
		&
		Three Faster R-CNNs (with Inception-ResNet-v2, Inception-v2, and ResNet-101 as the backbone networks, separately) and one SSD (with MobileNet-v2 as the backbone network) are trained to detect road potholes. Faster R-CNN (with ResNet-101 as the backbone network) achieves the best performance.
		\\
		
		Gupta {\etal} \cite{gupta2020detection} (2020) & Thermal image &
		SSD
		& Two SSDs (with ResNet-34 and ResNet-50 as the backbone networks, separately) are trained to detect potholes in thermal road images. The latter significantly outperforms the former.
		\\
		
		Javed {\etal} \cite{javed2021pothole} (2021) & Color image
		& R-CNN, SSD &
		R-CNN and SSD are compared on the road data collected in Bangladesh. They achieve similar road pothole detection performances.
		\\
		
		\bottomrule
	\end{tabular}
	\label{tab.object_detection_methods}
\end{table*}

Compared to SSD, R-CNN and YOLO series are more widely used for road pothole detection. In \cite{javed2021pothole}, R-CNN was demonstrated to achieve a similar performance to SSD for road pothole detection. In \cite{yebes2020learning}, four road pothole detection networks were developed: (1) Faster R-CNN (with Inception-v2 \cite{szegedy2016rethinking} as the backbone network), (2) Faster R-CNN (with ResNet-101 as the backbone network \cite{he2016deep}), (3) Faster R-CNN  (with Inception-ResNet-v2 as the backbone network \cite{szegedy2017inception}), and (4) SSD (with MobileNet-v2 \cite{sandler2018mobilenetv2} as the backbone network). Extensive experiments demonstrated that Faster R-CNN (with ResNet-101 as the backbone network) achieved the best overall performance. Their experimental results are shown in Fig. \ref{fig.faster_r-cnn}. \cite{wang2018road} compares the performance of two Faster R-CNNs (with ResNet-101 and ResNet-152 as the backbone networks, separately) for road damage detection on the dataset introduced in \cite{maeda2018road} w.r.t. three evaluation metrics: F1-Score, the harmonic mean of the precision, and the harmonic mean of the recall. The experimental results indicated that Faster R-CNN (with ResNet-152 as the backbone network) outperforms Faster R-CNN (with ResNet-101 as the backbone network). This is probably because a deeper backbone can learn more abstract representations. \cite{kortmann2020detecting} utilizes a Faster R-CNN to detect both cracks and potholes in the road images captured in Japan, India, and the Czech Republic. A classifier is first trained to infer the country where the road image was captured. A Faster R-CNN is then trained w.r.t. each country (in order to reduce the effects caused by the regional difference) for road crack and pothole detection.

\begin{figure*}[!t]
	\begin{center}
		\centering
		\includegraphics[width=0.99\textwidth]{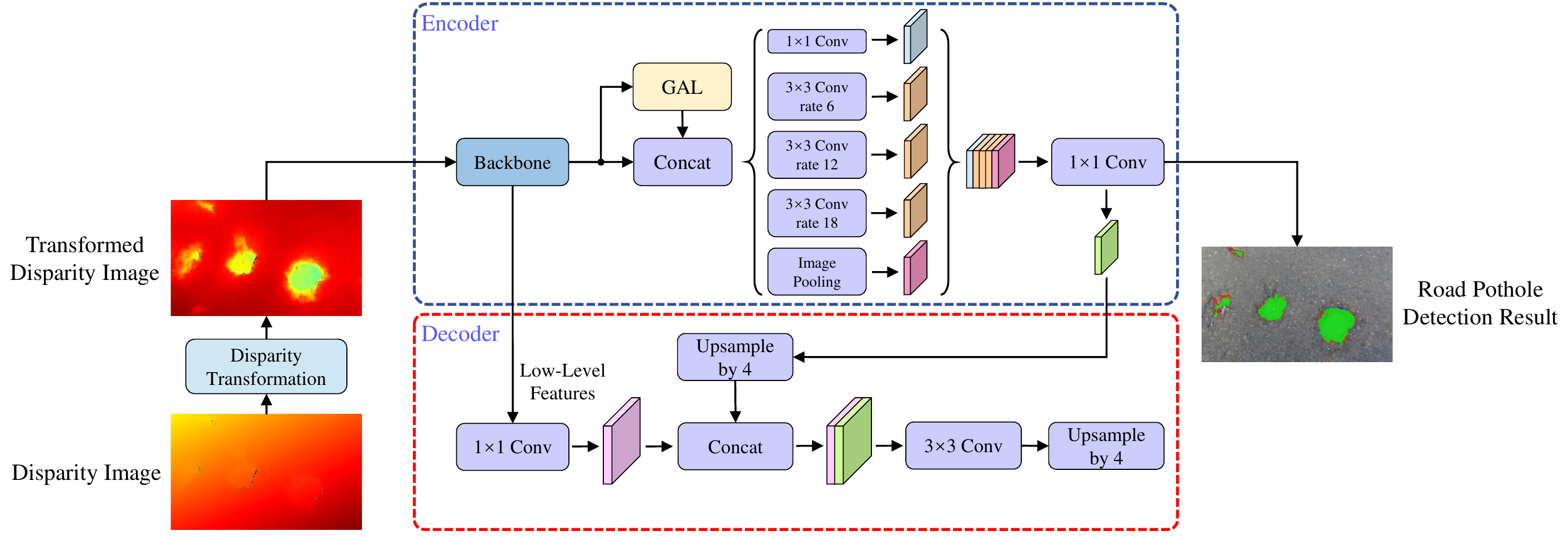}
		\centering
		\caption{Semantic segmentation for road pothole detection \cite{fan2021graph}. The disparity transformation algorithm was introduced in \cite{fan2019pothole,fan2019road}.}
		\label{fig.GAL-Deeplabv3+}
	\end{center}
\end{figure*}

\begin{table*}[!t]
	\renewcommand{\arraystretch}{1.3}
	\centering
	\caption{Semantic segmentation-based approaches. }
	\centering
	\begin{tabular}{C{2.0cm}C{2.2cm}C{3.0cm}p{8.7cm}}
		\toprule
		Reference & Input  & Key algorithm(s) & Details \\
		\hline  
		
		Pereira {\etal} \cite{pereira2019semantic} (2019) & Color image & U-Net &
		A conventional U-Net is trained to segment color road images for pothole detection. 
		\\
		
		Chun and Ryu \cite{chun2019road} (2019) & Color image & FCN & 
		An FCN is trained to segment color road images; A semi-supervised learning strategy is also employed to produce additional pseudo labels for network fine-tuning.
		\\
		
		Fan {\etal} \cite{fan2020we} (2020) & Color image, transformed disparity image & AA, GAN & 
		An AA framework and a training set augmentation technique are developed to enhance both single-modal and data-fusion semantic segmentation networks. The developed networks outperform all other SoTA networks.
		\\		
		Masihullah {\etal} \cite{masihullah2021attention} (2021) & Color image & DeepLabv3+ &
		An attention-based feature refinement module is incorporated into DeepLabv3+ for road pothole detection; The effectiveness of few-shot learning for road pothole detection is also validated. 		\\
		Fan {\etal} \cite{fan2021multi} (2021) & Color image, transformed disparity image & DeepLabv3+ &
		An MSFFM is proposed to refine the learning representations in single-modal semantic segmentation networks for road pothole detection.
		\\
		Fan {\etal} \cite{fan2021graph} (2021) & Color image, disparity image, transformed disparity image & DCNNs with GAL & 
		A GNN-inspired GAL is designed; GAL-DeepLabv3+ achieves the best road pothole detection performance over all other SoTA single-modal DCNNs on color images, disparity images, and transformed disparity images.
		\\
		\bottomrule
	\end{tabular}
	\label{tab.semantic_segmentation}
\end{table*}

Unlike the R-CNN series, which uses region proposals to localize road potholes within the image, the YOLO series generally splits the road image into a collection of grids, and within each grid, a collection of bounding boxes are selected. The network outputs a class probability and the offset values for each bounding box. The bounding boxes with the class probability above a threshold value are used to locate the road potholes within the image. Thanks to their accuracy and efficiency, the YOLO series has become the first choice for object detection-based road pothole detection. For example, in \cite{suong2018detection}, two object detection DCNNs, referred to as F2-Anchor and Den-F2-Anchor, were developed for road pothole detection. F2-Anchor, a variant of YOLOv2, is capable of generating five new anchor boxes (obtained using the K-means clustering algorithm \cite{hartigan1979algorithm}). The experimental results suggest that F2-Anchor outperforms the original YOLOv2 in detecting road potholes having different shapes and sizes. Compared to F2-Anchor, Den-F2-Anchor densifies the grid and achieves better road pothole detection performance than YOLOv2 and F2-Anchor. Additionally, \cite{dharneeshkar2020deep} trained three YOLO architectures: YOLOv3 \cite{redmon2018yolov3}, YOLOv2 \cite{redmon2017yolo9000}, and YOLOv3 tiny \cite{redmon2018yolov3}, for road pothole detection. YOLOv3 tiny achieved the best overall road pothole detection accuracy. Similarly, \cite{ukhwah2019asphalt} compared three different YOLOv3 architectures: YOLOv3 \cite{redmon2018yolov3}, YOLOv3 Tiny \cite{redmon2018yolov3}, and YOLOv3 SPP \cite{redmon2018yolov3}, for road pothole detection. YOLOv3 SPP demonstrated the highest road pothole detection accuracy. Recently, \cite{baek2020pothole} designed a hierarchical road pothole detection approach with two YOLOv1 networks \cite{redmon2016you}. One pre-trained YOLOv1 model was used to detect cars (background), while the other YOLOv1 was used to detect road potholes from the foreground. Nevertheless, the aforementioned object detection approaches can only recognize road potholes at the instance level, and they are infeasible when pixel-level road pothole detection results are desired.

\subsubsection{Semantic Segmentation-Based Methods}\label{sec.semantic_segmentation}

As shown in Fig. \ref{fig.mind_map}, the SoTA semantic segmentation networks are grouped into two major categories: (1) single-modal and (2) data-fusion. Single-modal networks generally segment RGB images with encoder-decoder architectures \cite{fan2021multi}. Data-fusion networks typically learn visual features from two different types of vision sensor data (color images and depth maps were used in FuseNet \cite{hazirbas2016fusenet}, color images and surface normal maps were used in SNE-RoadSeg series \cite{fan2020sne-roadseg,wang2021sneroadsegplus}, and color images and transformed disparity images were used in AA-RTFNet \cite{fan2020we}) and fuse the learned visual features to provide a better semantic understanding of the environment. The most representative prior arts are summarized in Table \ref{tab.semantic_segmentation}. 

\cite{chun2019road} proposes a road pothole detection approach based on fully convolutional network (FCN). To mitigate the difficulty in providing pixel-level annotations required by supervised learning, \cite{chun2019road} exploits a semi-supervised learning technique to generate pseudo labels and fine-tune the pre-trained FCN automatically. Compared to supervised learning, semi-supervised learning can greatly improve the overall F-score. Additionally, \cite{fan2021multi} incorporates an attention-based multi-scale feature fusion module (MSFFM) into DeepLabv3+ \cite{chen2018encoder} for road pothole detection. Similarly, \cite{masihullah2021attention} proposes an attention-based coupled framework for road pothole detection. This framework leverages an attention-based feature fusion module to improve the image segmentation performance. The work also demonstrates the effectiveness of few-shot learning for road pothole detection. 

We have conducted extensive research in this field. \cite{fan2020we} introduces an attention aggregation (AA) framework, which takes the advantages of three types of attention modules: (1) channel attention module (CAM), (2) position attention module (PAM), and (3) dual attention module (DAM). Additionally, \cite{fan2020we} develops an effective training set augmentation technique based on generative adversarial network (GAN), where fake color road images and transformed road disparity images are generated to enhance the training of semantic segmentation networks. The experimental results demonstrated that (1) AA-UNet (single-modal network) outperforms all other SoTA single-modal for road pothole detection, (2) AA-RTFNet (data-fusion network) outperforms all other SoTA data-fusion networks for road pothole detection, and (3) the training set augmentation technique not only improves the accuracy of the SoTA semantic segmentation networks but also accelerates their convergence during training. Recently, we developed a graph attention layer (GAL) based on graph neural network (GNN) to further optimize image feature representations for single-modal semantic segmentation \cite{fan2021graph}. As illustrated in Fig. \ref{fig.GAL-Deeplabv3+}, GAL-DeepLabv3+, the best performing implementation, outperforms all other SoTA single-modal semantic segmentation DCNNs for road pothole detection.  

It should be noted here that road pothole detection can be jointly solved with other driving scene understanding problems, notably freespace and road anomaly detection \cite{fan2020sne-roadseg, wang2020applying, wang2021dynamic, wang2021sneroadsegplus, fan2021learning}. Unfortunately, the SoTA semantic segmentation networks are strong data-driven algorithms that require a considerable amount of data. Therefore, road pothole detection with unsupervised or self-supervised learning is a popular area of research that requires more attention.

\subsection{Hybrid Methods}

Hybrid road pothole detection approaches typically leverage at least two categories of algorithms mentioned above, as shown in Fig. \ref{fig.mind_map}. They have been extensively studied for over a decade. Such approaches, as summarized in Tables \ref{tab.hybrid_2d_3d}, \ref{tab.hybrid_2d_ml}, and \ref{tab.hybrid_ml_3d}, have brought the SoTA results to this task. 

A decade ago, \cite{joubert2011pothole} developed a hybrid road pothole detection approach based on classical 2-D image processing as well as 3-D point cloud modeling and segmentation. An image gradient filter was first performed on the road videos (collected by a high-speed camera) to select keyframes that were considered to contain road potholes. The keyframes' 3-D road point clouds (acquired by Microsoft Kinect) were simultaneously modeled as planar surfaces. Similar to \cite{fan2018novel},  RANSAC was employed to enhance the robustness of 3-D road point cloud modeling. Road potholes were then detected by comparing the observed and modeled road surfaces. Thanks to the efficient 2-D image processing-based keyframe selection, the approach greatly reduces the redundant computations in 3-D point cloud modeling.
\cite{jog2012pothole} presents a similar hybrid approach. A road video collected by a high-definition camera is first processed to recognize the keyframes potentially containing road potholes. Simultaneously, this road video is also utilized for sparse-to-dense 3-D road geometry reconstruction. The road potholes are efficiently and accurately detected by analyzing such multi-modal road data. Such a hybrid method significantly reduces the number of incorrectly detected road potholes.  
\cite{jahanshahi2013unsupervised} introduces a similar hybrid road pothole detection approach based on RGB-D data (collected by a Microsoft Kinect) analysis. A planar surface is first fitted to the acquired depth image. Similar to \cite{joubert2011pothole}, this process is optimized with the RANSAC. A normalized depth-difference image, reflecting the difference between the actual and fitted depth images, is subsequently created and normalized. Otsu’s thresholding method is then performed on the normalized depth-difference image to detect road potholes. Recently, \cite{fan2019pothole} introduced a hybrid road pothole detection algorithm based on 2-D road disparity image transformation and 3-D road point cloud segmentation. A dense subpixel disparity map is first transformed to better distinguish between damaged and undamaged road areas. Otsu’s thresholding method is then used to extract potential undamaged road areas from the transformed disparity map. The disparities in the extracted regions are modeled as a quadratic surface using least-squares fitting (also improved with RANSAC). Surface normal information is also integrated into the point cloud modeling process to reduce outliers. Finally, the road potholes are effectively detected by comparing the actual and the modeled disparity maps. 

\begin{table*}[!t]
	\renewcommand{\arraystretch}{1.3}
	\caption{Hybrid approaches.}
	\centering
	\begin{subtable}[h]{1\textwidth}
		\caption{Classical 2-D image processing \& 3-D point cloud modeling and segmentation.}
		\centering
		\begin{tabular}{C{2.5cm}C{2.2cm}p{11.6cm}}
			\toprule
			Reference & Input & Details \\
			\hline
			Joubert {\etal} \cite{joubert2011pothole} (2011) & 3-D point cloud, color image &
			
			The keyframes (potentially containing road potholes) are selected using 2-D image processing algorithms; Road potholes in the keyframes are detected by comparing the observed and modeled 3-D road point clouds.
			\\
			
			Jog {\etal} \cite{jog2012pothole} (2012) & 3-D point cloud, color image &
			The road videos are analyzed with 2-D image processing algorithms to produce keyframes; The road videos are also used to reconstruct 3-D road geometry for road pothole detection. 
			\\
			
			Jahanshahi {\etal} \cite{jahanshahi2013unsupervised} (2013) & 3-D point cloud, depth image &
			A planar surface is fitted to the depth image; A normalized depth-difference image, reflecting the difference between the observed the fitted depth images, is created; 
			Otsu’s thresholding method is used to segment the normalized depth-difference image for road pothole detection. 
			\\
			
			Fan {\etal} \cite{fan2019pothole} (2019) & 3-D point cloud, transformed disparity image &
			A disparity image is transformed into a quasi bird's eye view. Otsu’s thresholding method is utilized to segment the transformed disparity image to produce the undamaged road areas. The 3-D points in the undamaged road areas are used to interpolate a quadratic surface; Road potholes are detected by comparing the observed and interpolated surfaces. 
			\\
			\bottomrule
		\end{tabular}
		\label{tab.hybrid_2d_3d}
	\end{subtable}
\end{table*}

\begin{table*}[!t]
	\renewcommand{\arraystretch}{1.3}
	\ContinuedFloat
	\centering
	\begin{subtable}[h]{1\textwidth}
		\caption{Classical 2-D image processing \& machine/deep learning.}
		\centering
		\begin{tabular}{C{2.5cm}C{2.2cm}p{11.6cm}}
			\toprule
			Reference & Input & Details\\
			\hline
			
			Azhar {\etal} \cite{azhar2016computer} (2016) & Color image & 
			HOG features are extracted from road images; An NBC is trained with the HOG features to classify road images. The NGCS method is used to segment road images potentially containing potholes.
			\\
			
			Yousaf {\etal} \cite{yousaf2018visual} (2018) & Color image &
			Road images are classified using the BoW algorithm; The GCS is used to segment the road images that potentially contain potholes.
			\\
			
			Anand {\etal} \cite{anand2018crack} (2018) & Color image
			&
			A SegNet is trained to segment road images for freespace detection; The freespace regions are processed to generate road pothole/crack candidates; A SqueezeNet is trained to determine whether the generated candidates were road potholes or cracks.
			\\
			\bottomrule
		\end{tabular}
		\label{tab.hybrid_2d_ml}
	\end{subtable}
\end{table*}
\begin{table*}[!t]
	\renewcommand{\arraystretch}{1.3}
	\ContinuedFloat
	\centering
	\begin{subtable}[h]{1\textwidth}
		\caption{Machine/deep learning \& 3-D point cloud modeling and segmentation.}
		\centering
		\begin{tabular}{C{2.5cm}C{2.2cm}p{11.6cm}}
			\toprule
			Reference & Input & Details \\
			\hline
			
			Dhiman and Klette \cite{dhiman2019pothole} (2019) & 3-D point cloud, color image, disparity image & 
			
			Four existing computer vision techniques are compared: (1) single-frame stereo vision-based method; (2) multi-frame vision sensor data fusion-based method; (3) Mask R-CNN  trained with transfer learning; and (4) YOLOv2 trained with transfer learning.
			\\
			
			Wu {\etal} \cite{wu2019road} (2019) & 3-D point cloud, color image &
			A semantic segmentation network is used to provide initial road pothole detection results; A 3-D point cloud modeling and segmentation algorithm is used to refine such results and calculate the road pothole volumes.  
			\\
			\bottomrule
		\end{tabular}
		\label{tab.hybrid_ml_3d}
	\end{subtable}
\end{table*}

In addition to the approaches discussed above, researchers  developed hybrid approaches based on classical 2-D image processing algorithms and machine/deep learning models. Taking \cite{azhar2016computer} as an example, a naive Bayes classifier (NBC) \cite{rish2001empirical} is trained to learn histograms of oriented gradients (HOG) \cite{dalal2005histograms} features. Such HOG features are then utilized to train a road image classifier. Once an image is considered to contain road potholes, it is segmented using the normalized graph cut segmentation (NGCS) \cite{shi2000normalized} algorithm to produce a pixel-level road pothole detection result. Furthermore, \cite{yousaf2018visual} proposes a two-stage road pothole detection approach. In the first stage, the bag of words (BoW) \cite{csurka2004visual} algorithm is utilized to classify road images. This process has four steps: (1) scale-invariant feature transform (SIFT) \cite{lowe2004distinctive} feature extraction and description, (2) visual vocabulary/codebook construction with K-means clustering, (3) histogram of words generation, and (4) road image classification with SVM. In the second stage, the graph cut segmentation (GCS) \cite{shi2000normalized} algorithm is used to segment road images for pixel-level road pothole detection. Recently, \cite{anand2018crack} proposed a hybrid road crack and pothole detection algorithm. A modified SegNet \cite{badrinarayanan2017segnet} is first trained to segment road images for freespace detection. The freespace regions are then processed with a Canny edge detector to generate road crack/pothole candidates. Finally, a SqueezeNet \cite{iandola2016squeezenet} is trained to determine whether the generated candidates are road cracks or potholes.

In recent years, road pothole detection approaches based on 3-D point cloud segmentation and machine/deep learning have also attracted much attention.  \cite{dhiman2019pothole} is a representative prior art in this field. \cite{dhiman2019pothole} compares four existing computer vision techniques for road pothole detection: (1) SV1, a single-frame stereo vision-based method, based on $v$-disparity image analysis and 3-D plane fitting (in disparity space); (2) SV2, a multi-frame vision sensor data fusion-based method, developed based on the digital elevation model (DEM) and visual odometry; (3) LM1, Mask R-CNN \cite{he2017mask} trained with transfer learning; and (4) LM2, YOLOv2 \cite{redmon2017yolo9000} trained with transfer learning. 
Furthermore, \cite{wu2019road} introduced a hybrid road pothole detection method based on semantic road image segmentation and 3-D road point cloud segmentation. 
A DeepLabv3+ \cite{chen2018encoder} model is first trained to produce initial pixel-level road pothole detection results. The 3-D points of the initially detected road potholes' edges are classified as exterior and interior ones. The exterior edges are used to fit local planes and calculate road pothole volumes, while the interior edges are used to reduce incorrectly detected potholes by analyzing the road depth distribution.

\section{Public Datasets}\label{sec.public_datases}
This section briefly introduces the existing open-access road pothole detection datasets, which can provide researchers with indications of appropriate datasets when evaluating their developed road pothole detection algorithms.

\cite{viren} created a dataset for road image classification. It consists of a training set and a test set. The training set contains 367 color images of healthy roads and 357 color images of roads with potholes; The test set contains eight color images of each category. This dataset is available at \url{kaggle.com/virenbr11/pothole-and-plain-rode-images}.

\cite{rath} presented a large-scale dataset for instance-level pothole detection. This dataset consists of a training set, a test set, and an annotation CSV file. The training set contains 2,658 color images of healthy roads and 1,119 color images of roads with potholes. The test set contains 628 color images. The images (resolution: $2760\times3680$ pixels) were captured using a GoPro Hero 3+ camera. This dataset can be accessed at \url{kaggle.com/sovitrath/road-pothole-images-for-pothole-detection}.

\cite{Yantra} created a dataset (image resolution: $720\times1280$ pixels) of Indian roads, with semantic segmentation annotations (road, pothole, footpath, shallow path, and background). This dataset contains a training set of 2,475 color images and a test set of 752 color images. This dataset is available at \url{kaggle.com/eyantraiit/semantic-segmentation-datasets-of-indian-roads}.

\cite{guzman2015towards} created a dataset, referred to as CIMAT Challenging Sequences for Autonomous Driving (CCSAD). It was initially created to develop and test autonomous vehicle perception and navigation algorithms. The CCSAD dataset includes four scenarios: (1) colonial town streets, (2) urban streets, (3) avenues and small roads, and (4) a tunnel network. This dataset contains 500 GB of high-resolution stereo images, complemented with inertial measurement unit (IMU) and GPS data. The CCSAD dataset is publicly available at \url{aplicaciones.cimat.mx/Personal/jbhayet/research}. 

\cite{maeda2018road} presented a large-scale road damage dataset, including 9,053 color road images (resolution: 600$\times$600 pixels) collected in Japan.
The images (containing 15,435 road damages) were captured using a smartphone mounted on a car under different weather and illumination conditions.
This dataset is publicly available at \url{github.com/sekilab/RoadDamageDetector}.

\cite{Chitholian} created a dataset of 665 pairs of color road images and pothole ground truth labels under different road conditions. This dataset can be used for automatic pothole detection and localization in urban streets. This dataset is publicly available at \url{public.roboflow.com/object-detection/pothole}.

Another road pothole detection dataset \cite{Kumar} was created for binary road image classification. It contains 352 undamaged road images and 329 pothole images. This dataset is small and can only be used to test image classification CNNs. It is available at \url{kaggle.com/datasets/atulyakumar98/pothole-detection-dataset}.

\cite{fan2019pothole} published the world’s first multi-modal road pothole detection dataset (image resolution: $800\times1312$ pixels), containing 55 groups of (1) color images, (2) subpixel disparity images, (3) transformed disparity images, and (4) pixel-level pothole annotations.
This dataset is publicly available at \url{github.com/ruirangerfan/stereo\_pothole\_datasets}.

Pothole-600 \cite{fan2020we} was recently published by the same research group.
It also provides two modalities of vision sensor data: (1) color images and (2) transformed disparity images. 
The transformed disparity images were obtained by performing the disparity transformation algorithm \cite{fan2018novel} on dense subpixel disparity images estimated using the stereo matching algorithm introduced in \cite{fan2018road}.
The Pothole-600 dataset is available at \url{sites.google.com/view/pothole-600}.

\section{Existing Challenges and Future Trends}\label{sec.existing_challenges}
Before the deep learning boom in 2012, classical 2-D image processing-based approaches dominated this research field. Such explicit programming approaches are, however, usually computationally intensive and sensitive to various environmental factors, most notably illumination and weather conditions \cite{jahanshahi2013unsupervised}. Furthermore, road potholes have irregular shapes, making the geometric assumptions made in such approaches occasionally infeasible. Therefore, since 2013, 3-D point cloud modeling and segmentation-based approaches have emerged to boost the road pothole detection accuracy \cite{zhang2013advanced}. Nevertheless, such approaches generally require a small field of view because of the assumption that a single-frame 3-D road point cloud is a planar or quadratic surface. Although significant efforts have been made to further improve the robustness of road point cloud modeling, such as using the RANSAC algorithm \cite{fan2019pothole}, extensive parameters are required to ensure the satisfactory performance of these approaches, making them highly challenging to adapt to a new scenario.

Over the past five years, DCNNs have been widely used to solve this problem. Image classification networks can only determine whether a road image contains potholes. Object detection networks can only provide instance-level road pothole detection results. Since the transportation departments are more concerned about potholes' geometric properties, such as width, depth, volume, \etc., developing hybrid approaches that combine 3-D road geometry reconstruction and semantic segmentation is the future trend of this research. 

Recent deep stereo matching networks have demonstrated superior performance. We believe that they can be easily applied to reconstruct 3-D road geometry models through transfer learning. However, such (supervised) approaches typically require a large amount of well-labeled training data to learn stereo matching, making them often hard to implement in practice \cite{wang2021pvstereo}. Therefore, un/self-supervised stereo matching algorithms, specifically developed for road surface 3-D reconstruction, are also a popular research area that requires more attention. Furthermore, as stated in \cite{fan2020sne-roadseg, wang2021sneroadsegplus, wang2020applying, wang2021dynamic}, data-fusion semantic segmentation is currently a hot topic in driving scene understanding. However, such networks are generally computationally complicated. After extensive literature investigation, we believe that network pruning and knowledge distillation can be feasible solutions to this problem. In practical experiments, we can also apply a well-trained image classification DCNN to select keyframes (the road images that potentially contain potholes), significantly avoiding the redundant computations of semantic segmentation. Road potholes are not necessarily ubiquitous, and it is challenging to prepare a large, well-annotated dataset to train semantic segmentation DCNNs. Therefore, developing few/low-shot semantic segmentation networks for road pothole detection is also a popular area of research that requires more attention.

\section{Conclusion}\label{sec.conclusion}

This article comprehensively reviewed the SoTA road imaging techniques and computer vision algorithms developed for road pothole detection. Classical 2-D image processing-based and 3-D point cloud modeling and segmentation-based approaches have serious limitations. Hence, this article mainly discussed the well-performing SoTA DCNNs, developed for road pothole detection. Since transportation departments are more interested in the  geometric properties of potholes, developing hybrid approaches, consisting of stereo matching-based road surface 3-D reconstruction and data-fusion semantic segmentation functionalities, is the future trend of this research. However, training stereo matching and semantic segmentation networks requires large human-annotated datasets, and preparing such datasets is exceptionally labor-intensive. Therefore, we believe that un/self-supervised stereo matching algorithms, developed specifically for road surface 3-D, and few/low-shot learning for semantic road image segmentation are popular areas of research that require more attention.

\section*{Acknowledgment}
This work was supported by the National Key R\&D Program of China (Grant no. 2020AAA0108100).

\end{document}